\documentclass[10pt, twocolumn]{article}   	
\usepackage{geometry}                		
\pdfoutput=1                 		
\usepackage{graphicx}
\usepackage{subfigure}				
\usepackage{amssymb}
\usepackage{amsmath}
\usepackage{url}
\title{\bf CT Image Registration in Acute Stroke Monitoring}
\author{Lucio Amelio$^1$\footnote{(Corresponding author)}, Alessia Amelio$^2$}
\date{\small $^1$  Undergraduate, Faculty of Medicine and Surgery, University of Bologna, Italy\\ lucio.amelio@studio.unibo.it\\$^2$Department of Computer Engineering, Modeling, Electronics and Systems, University of Calabria, Italy\\aamelio@dimes.unical.it}							

\begin{document}
\maketitle

\begin{@twocolumnfalse}

{\bf Abstract}. We present a new system based on tracking the temporal evolution of stroke lesions using an image registration technique on CT exams of the patient's brain. The system is able to compare past CT exams with the most recent one related to stroke event in order to evaluate past lesions which are not related to stroke. Then, it can compare recent CT exams related to the current stroke for assessing the evolution of the lesion over time. A new similarity measure is also introduced for the comparison of the source and target images during image registration. It will result in a cheaper, faster and more accessible evaluation of the acute phase of the stroke overcoming the current limitations of the proposed systems in the state-of-the-art.
\end{@twocolumnfalse}
\vspace{1cm}

\section{Introduction}
\begin{figure}[t!]
\begin{center}
\includegraphics[height=7cm, width=7cm, keepaspectratio]{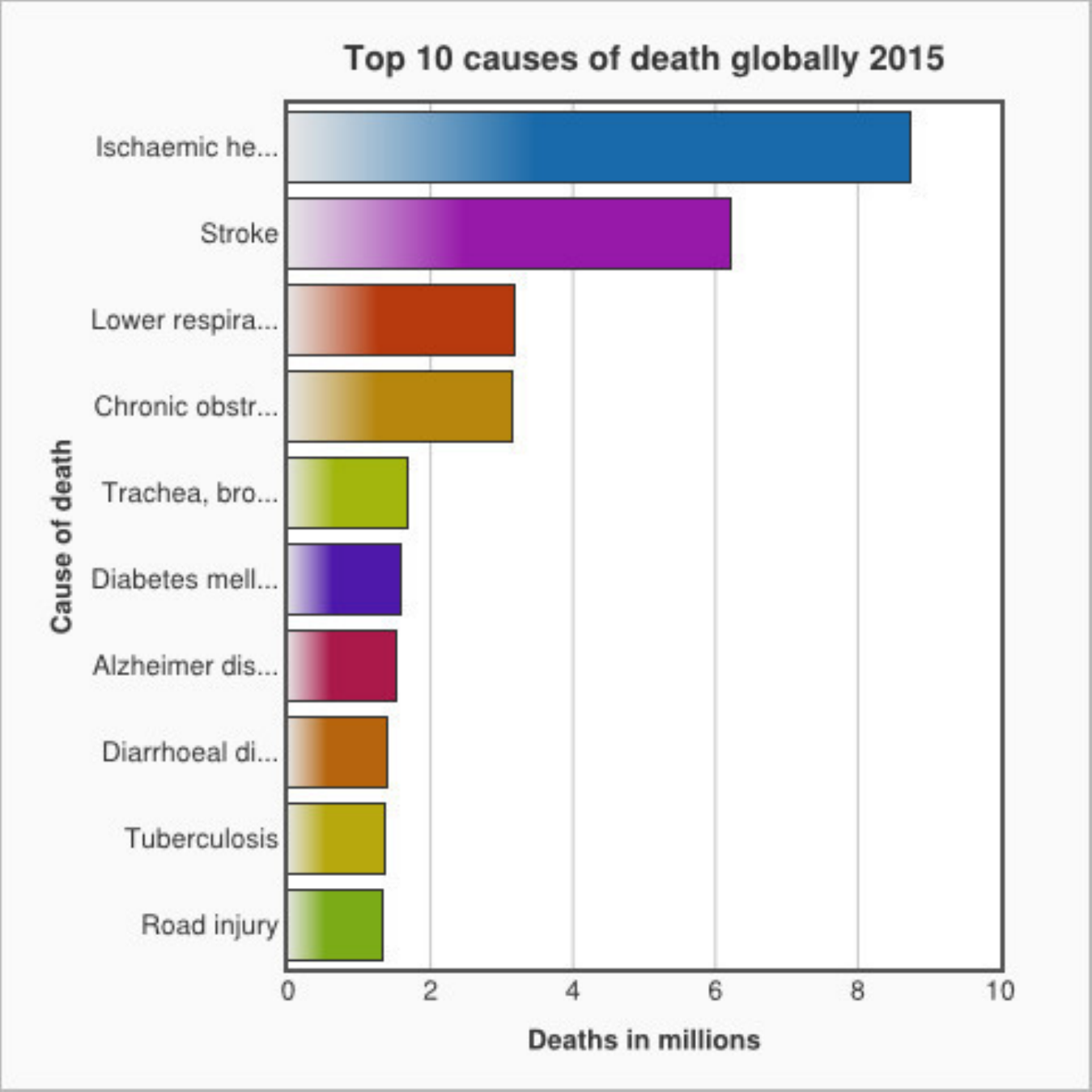}
\caption{Histogram of the top 10 causes of death worldwide in 2015. Stroke is the second one in the order}
\label{Fig1}
\end{center}
\end{figure}

Cardiovascular diseases (CVD) are today the main cause of death in developed countries (Europe and USA) and in some developing ones. According to World Health Organization (WHO), on 56.4 millions of deaths worldwide during 2015, about 15 millions were directly caused by CVD whose stroke represents about 6 millions. Fig. \ref{Fig1} shows the distribution of the main death causes worldwide during 2015. It is worth noting that stroke is at second position.

\begin{figure}[t!]
\begin{center}\fbox{
\includegraphics[height=7cm, width=7cm, keepaspectratio]{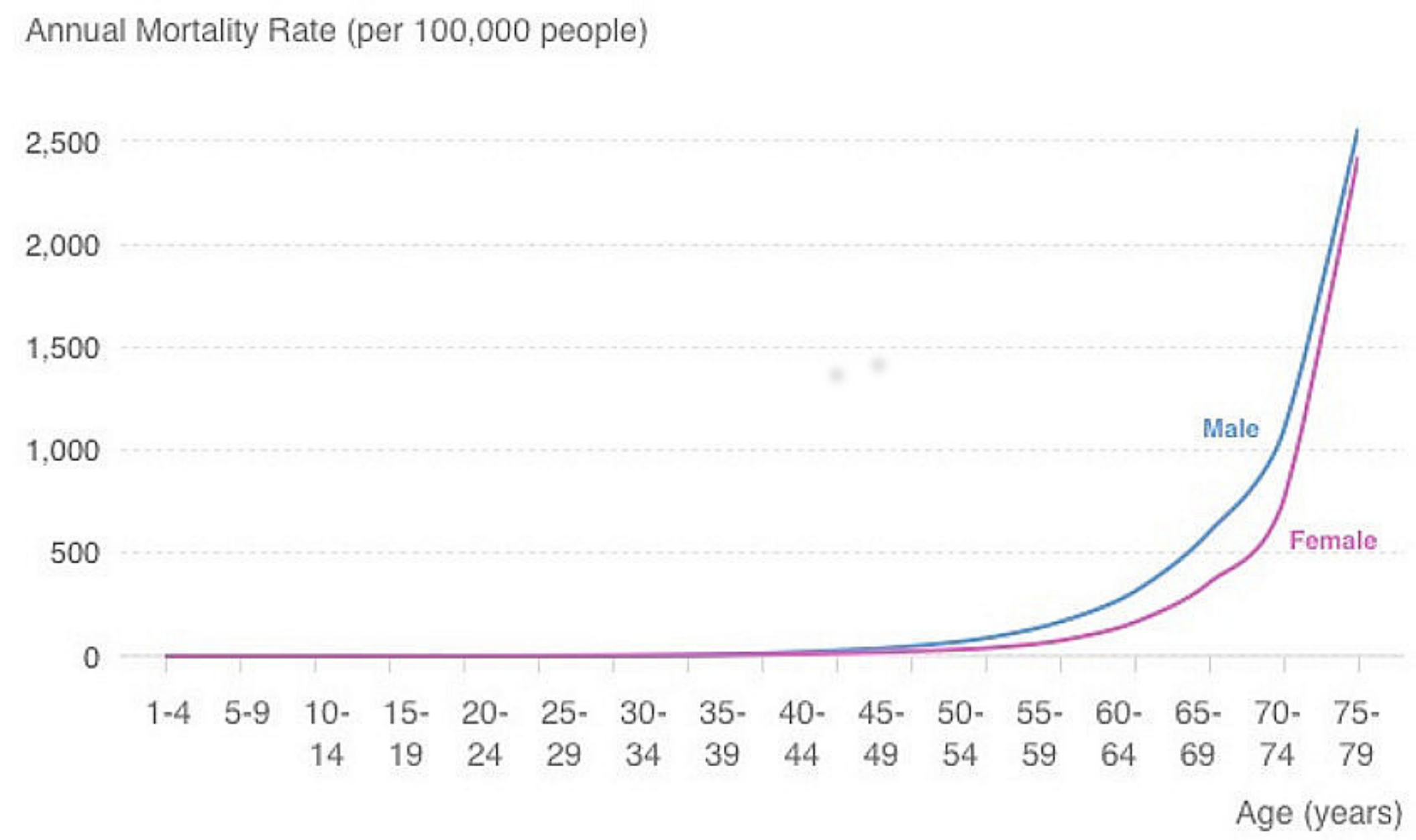}}
\caption{Annual mortality rate in Croatia during 2013}
\label{Fig2}
\end{center}
\end{figure}

In Croatia, the annual mortality rate for stroke reached 197.4 per 100'000 people. Fig. \ref{Fig2} shows the trend of annual mortality rate over age in Croatia during 2013. The deadliness of stroke exhibited the lowest value for male subjects of 5-9 years old and rapidly increased with age. In 2013, the highest peak was reached at 75-79 years range with 2'559.7 deaths per 100'000 men and 2'424.3 per 100'000 women. It is worth noting that male subjects are more exposed to death risk for stroke than female ones. 
Consequently, in Europe as well as in the rest of the world the stroke has represented and still represents a serious problem which needs to be taken into high consideration. For this reason, it is of great importance to aim resources in this direction in terms of research and development.
The general approach to stroke requires an immediate assistance in emergency units located in hospitals where the first medical aids are applied, including a first medical examination and other specific exams. After the hyperacute phase, the patient is moved to a stroke unit (if the gravity level is not too low or high) where further advanced aids are performed from an internal medicine point of view to neurological and cardiological ones. In the entire phase of the patient management, the role of medical imaging is essential to support the physician during all the steps of diagnosis and treatment decisions. During the days after the event (acute phase), it is really important to monitor the evolution of patientÕs conditions. It can be successfully accomplished by following the eventual evolution of encephalic lesions or the appearance of new ones using techniques of medical imaging.

\subsection{Computed Tomography imaging}

\begin{figure}[t!]
\begin{center}
\includegraphics[height=5.5cm, width=5.5cm, keepaspectratio]{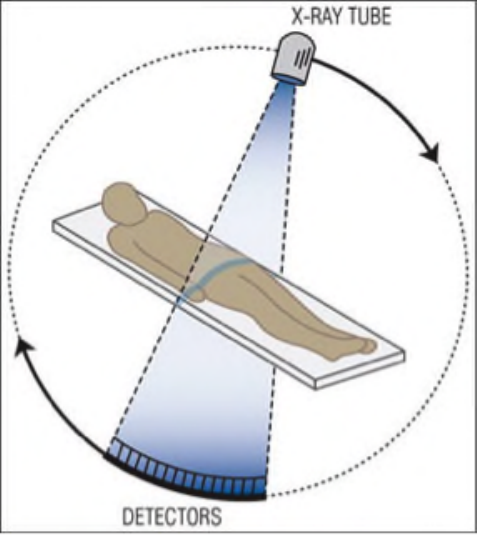}
\caption{CT imaging procedure: x-ray source and detectors}
\label{Fig3}
\end{center}
\end{figure}

Computed Tomography (CT) imaging is a type of medical imaging which generates a 3D view of internal body parts using x-ray and computer-aided tomographic imaging methods \cite{[1]}. In particular, the patient is exposed to an x-rays beam. The projected image is captured on a semicircular x-ray detector. The patient is positioned between the x-ray source and the detector. Each detected image corresponds to an x-ray projection of a transverse slice of the body part. In order to generate the CT exam, the x-ray source and detector are rotated around the patient, and multiple images are generated and stored. Given an image depicting a single slice, each pixel represents the relative radio density of the patient along the line connecting the x-ray source to the pixel position. Fig. \ref{Fig3} shows a simplified procedure of capturing an image by a digital x-ray detector.

\begin{figure}[t!]
\begin{center}
\subfigure[]{
\includegraphics[height=3.2cm, width=3.2cm, keepaspectratio]{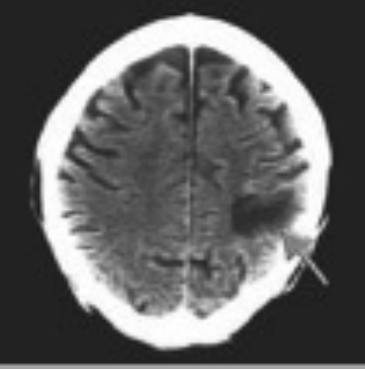}
}
\subfigure[]{
\includegraphics[height=3.2cm, width=3.2cm, keepaspectratio]{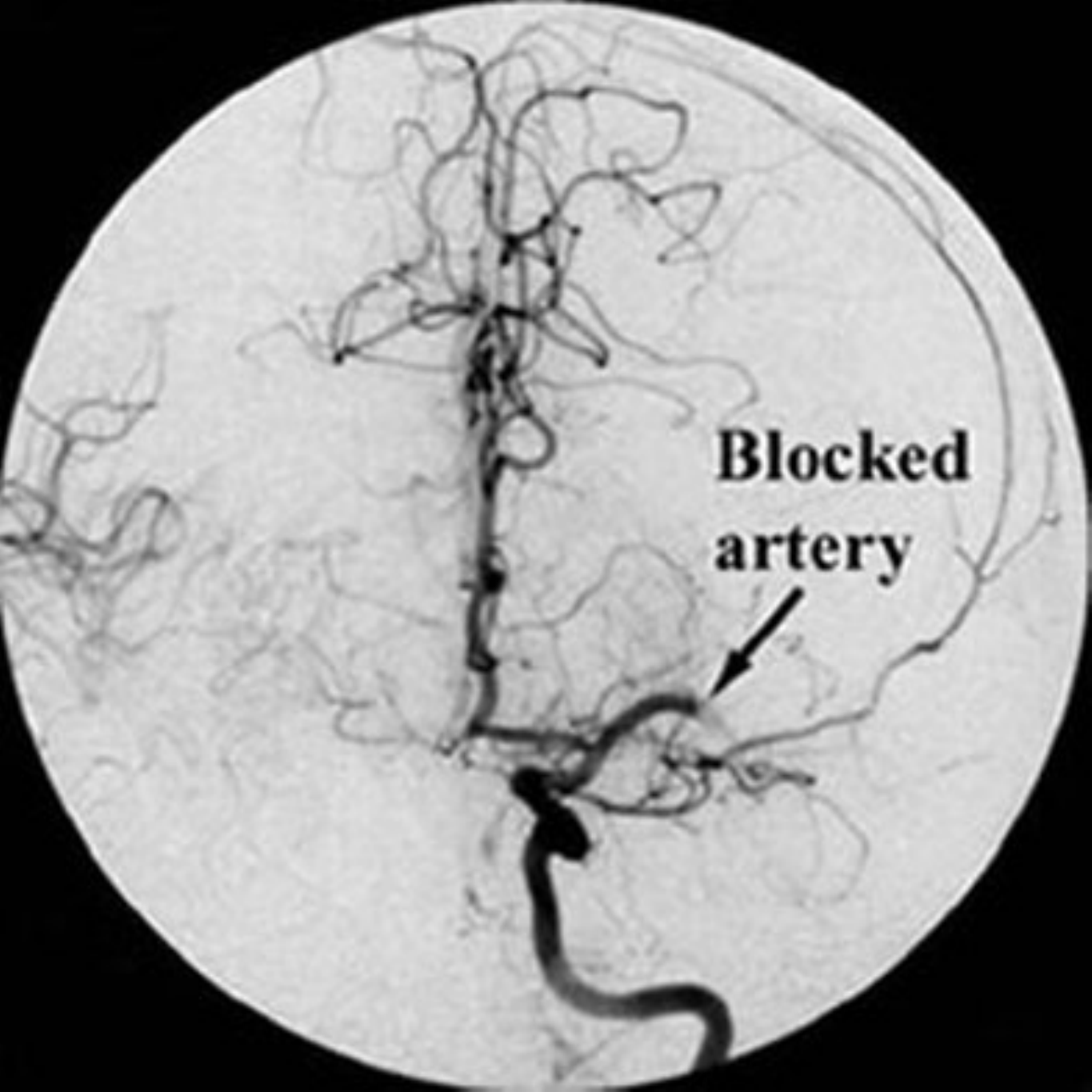}
}
\caption{CT image samples: (a) brain CT image with stroke (green arrow on the right), and (b) CT angiography image with blocked artery}
\label{Fig4}
\end{center}
\end{figure}

CT imaging is usually one of the most important and first exams which are generated for stroke evaluation. It is particularly useful in the hyperacute stroke phase in the emergency room \cite{[2]}. In fact, a non contrast encephalic CT exam is essential for capturing the early signs of stroke, and for excluding intracerebral hemorrhage and lesions which could be mistaken for an acute ischemic stroke \cite{[3]}. Also, a CT angiogram is particularly useful for monitoring the state of the blood vessels in order to identify possible occlusions, which supports the stroke diagnosis. In particular, it can be adopted for capturing eventual occlusions of the extracranical carotid arteries or even eventual occlusions of the blood vessels in the brain. Fig. \ref{Fig4} shows a brain CT image with stroke (see (a)), and a brain CT angiography image with occluded artery (see (b)).

\subsection{Related work}

In the literature, monitoring the temporal evolution of the stroke is still poorly explored and captured using image registration techniques applied on medical imaging. This is a method of computer vision finalized to compare 2D or 3D images by mapping specific coordinates in a space of an image to the corresponding ones in another space of another image in order to define an alignment. In particular,  the actually proposed image registration techniques for stroke analysis are used on MR (magnetic resonance) images and on CT  angiography images for the study of the carotid state. In the context of MR imaging, Ref. \cite{[4]} analysed the impact of nonlinear image registration in more accurately identifying the infarct on serial MR images captured at different time points than rigid body registration methods. Also, Ref. \cite{[5]} introduced a robust method of stroke image registration which registered the ischemic lesion on follow-up MRI with the first generated MRI. Dalca et al. \cite{[6]} proposed a new patch-based non-rigid image registration method of brain images which compared 3D patches in order to compute the similarity. It was tested on clinical MR images of stroke patients. Aganj and Fischl \cite{[7]} introduced a noninformation-theoretic image registration method applied on MR images which used the error of segmenting both images with a single segmentation as the registration cost function. Crum et al. \cite{[8]}  proposed an affine registration method representing the MR image population as a rooted-tree chain graph. Edges of the graph were weighted according to a distance function. The nearest neighbour images in the graph were first registered. On the contrary, the images more distant by one edge in the graph were registered by transformation composition. Finally, Ref. \cite{[9]} introduced an image processing system which included a phase of registration of brain MR images captured in different time points for the identification of the areas affected by the stroke. In the context of the CT angiogram, Ref. \cite{[10]} proposed a pre-processing step for CT stroke diagnosis based on 3D image registration of dynamic CT exams and CT angiograms. Also, Ref. \cite{[11]} introduced an image registration method for registering multi modal (CT or MR) brain exams and carotid arteries from ascending aorta to brain angiograms with multiple focus. 

There are different main limitations in the aforementioned methods of the state-of-the-art:
\begin{enumerate}
\item the high costs related to MR imaging, 
\item the low availability in different contexts of MR approaches, 
\item higher times for MR image acquisition, and
\item the acute phase temporal monitoring of the stroke is rarely in the focus.
\end{enumerate}

The paper is organised as follows. Section II describes our contribution and how it overcomes the current limitations in the state-of-the-art. Section III presents the proposed system, including the image registration method, the adopted similarity measure, and the image fusion method. Section IV discusses the expected results and benefits from the system. Finally, Section V concludes the paper and outlines future work directions.

\section{Our Contribution}
\begin{figure*}[t!]
\begin{center}
\includegraphics[height=9cm, width=9cm, keepaspectratio]{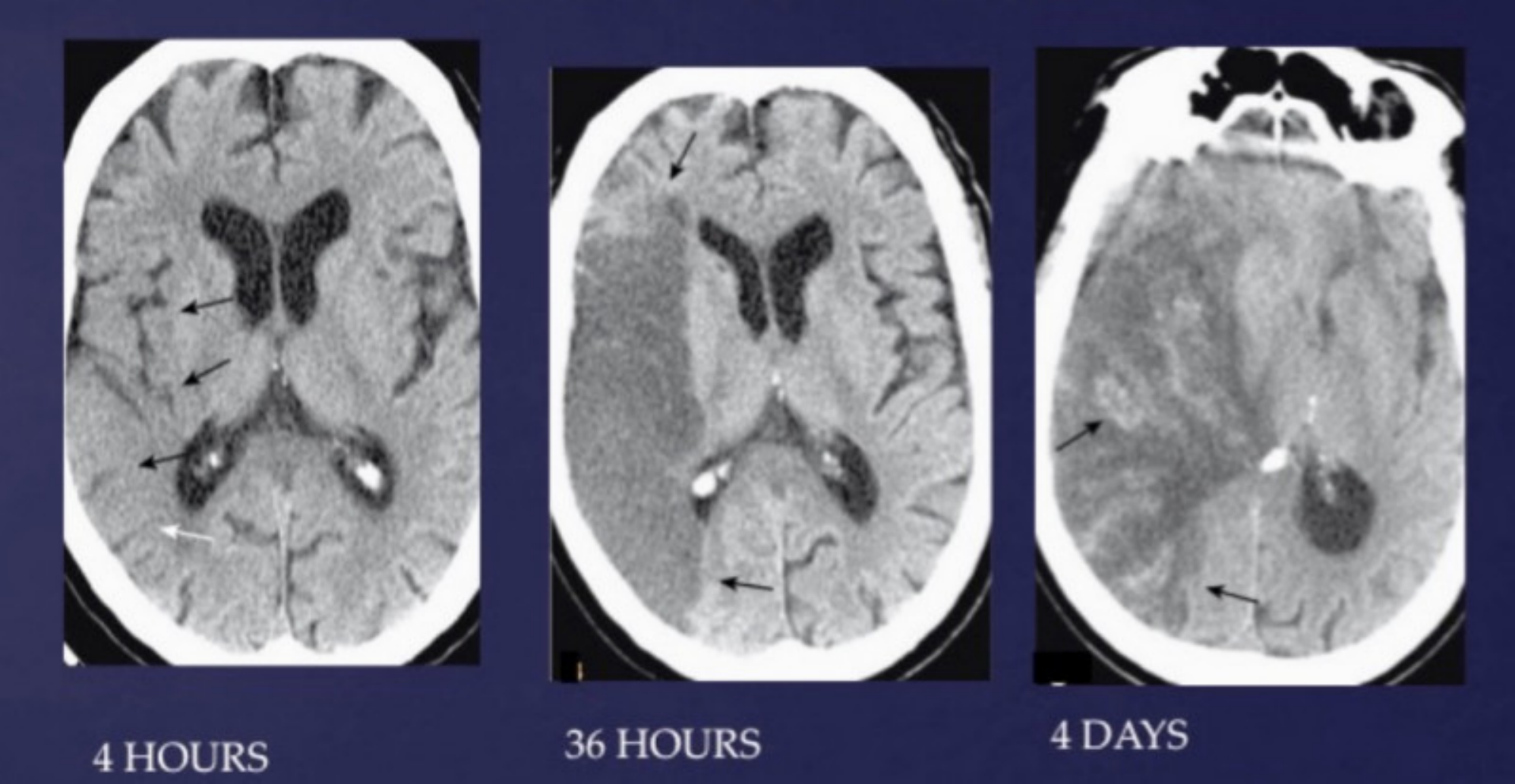}
\caption{Progression of stroke at various time references on CT scans \cite{[12]}: Radiology \& Imaging, Saveetha University, Saveetha Medical College Hospital - Chennai/IN}
\label{Fig5}
\end{center}
\end{figure*}

In this context, we propose a new automatic system that applies image registration techniques and subsequent image fusion to sequential encephalic CT exams in order to monitor the radiological patterns variation and determine the patientÕs situation step by step during the acute phase of the stroke. This will help the physician with the decision making process for a faster and more integrated management of the clinical state. In a first step, the system scans for patientÕs personal information in the hospital radiological database to identify eventual past CT exams which defines ``source images" to compare with the pathological as recent as possible one. This is finalised to take over eventual past lesions which are not related to current stroke. In a second step, the system only compares the most recent CT scans of the current stroke in order to evaluate the variation, over time, of the encephalic lesions. The system employs a new similarity measure for the comparison of two CT exams during the registration process. This system will result in a cheaper, faster and more accessible evaluation and monitoring of patient's state during the acute phase of the stroke. Fig. \ref{Fig5} shows how the evolution of a stroke at different time points is visible on CT scans.

\section{System Description}
\begin{figure*}[t!]
\begin{center}
\includegraphics[height=14cm, width=14cm, keepaspectratio]{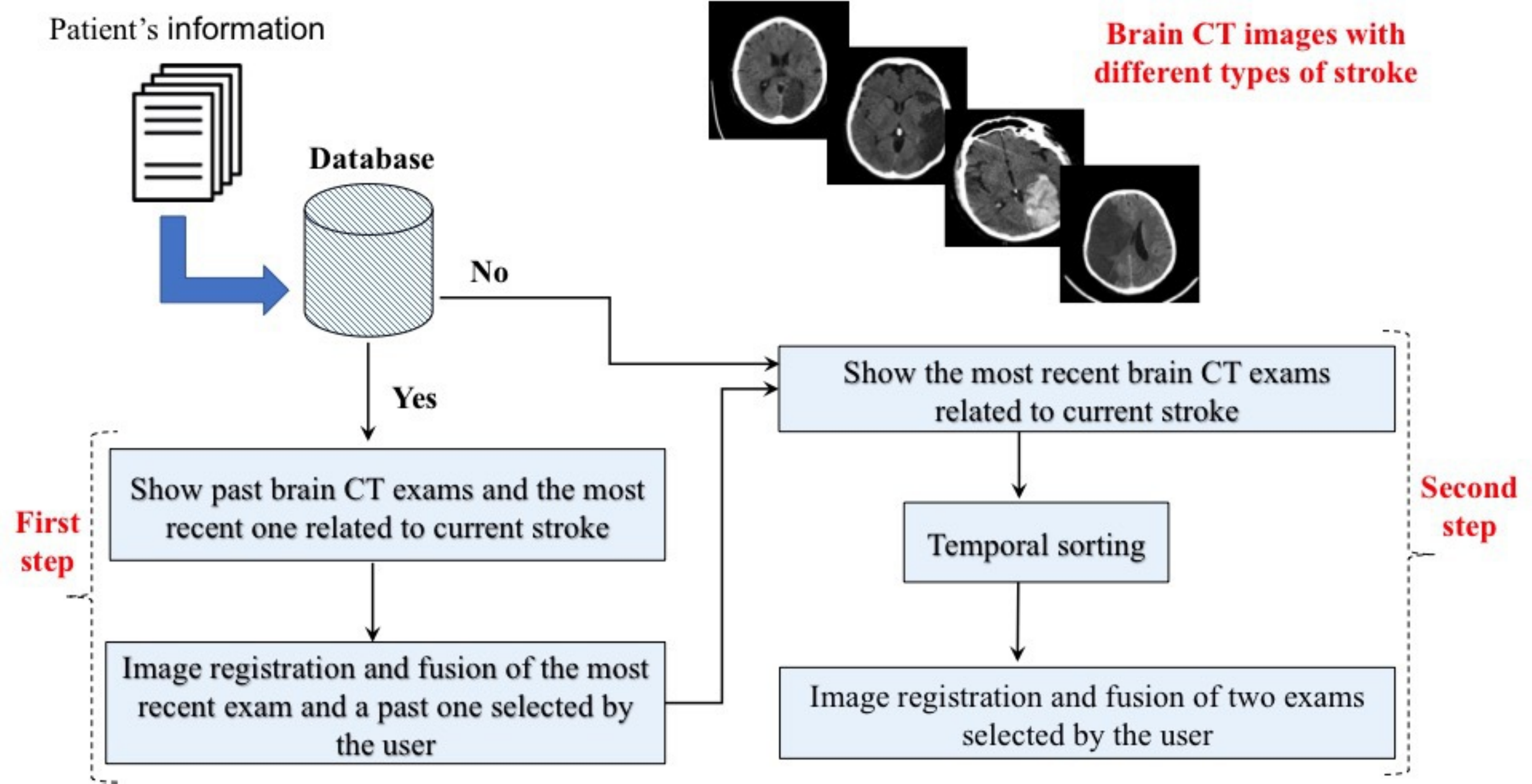}
\caption{Overview of the proposed system}
\label{Fig6}
\end{center}
\end{figure*}

The proposed system consists in two main steps, which are depicted in Fig. \ref{Fig6}. Initially, it takes the patient's information and checks for the presence into the radiological database. If data is found, the first step starts. In particular, CT exams of the patient are filtered and only brain CT ones are offered to the physician to be selected. The selection of a specific exam depends on the circumstances they were executed (diagnosis, date). After the exam is selected, it is compared with the most recent available brain CT exam related to stroke using image registration and image fusion. After the first step or if brain CT exams of the patient are not found in the database, the second step starts. In particular, the system takes the most recent brain CT exams related to the current stroke and temporally sorts them. After that, the system gives the possibility to select two exams to compare for assessing the stroke evolution during the acute phase using image registration and fusion.

\subsection{Image registration}

We employ an intensity-based image registration which calculates a transformation between two CT exams using the voxel values alone \cite{[13]}. The aim of the registration is to compute a transformation $T$ such that the selected similarity function $S$ is maximised between the source CT exam $A$ and a transformation of the target CT exam $T(B)$. The similarity measure is computed from the voxel values. Accordingly, the image registration problem is as follows:

\begin{equation}
T=arg \max_{T \in \textrm{T}} S(A, T(B)),
\end{equation}
 			
where T is a class of allowed transformations. Hence, the image registration is considered as an optimisation problem, where the target CT exam $B$ is iteratively mapped to $T(B)$ by progressively estimating the transformation $T$. For a faster approach, we employ the class of global and rigid transformations, which are very used in the medical context with good approximation in most cases. They can include translation, rotation, scaling, affine transformation, and perspective projections, such that they preserve distances, lines and angles. The rigid transformation $T$ is represented as \cite{[14]}:

\begin{equation}
\begin{bmatrix} 
x'  \\
y' \\
z'
\end{bmatrix}
=
\begin{bmatrix} 
w_{11} & w_{12} & w_{13} \\
w_{21} & w_{22} & w_{23} \\
w_{31} & w_{32} & w_{33}
\end{bmatrix}
\begin{bmatrix} 
x \\
y \\
z
\end{bmatrix},
\end{equation}\\

where $w_{ij}$, $i$,$j$=1,...,3 are the parameters of the transformation $T$.
Accordingly, we are searching for the parameters of the transformation $T$ which maximise the similarity. In order to solve the optimisation problem, the similarity function is evaluated for multiple parameters. 

Usually, only a subset of voxels is selected for the registration in order to speed-up the process. It corresponds to a sort of sampling of the CT exam in order to speed-up the registration. We do not provide any prior sampling of the CT exams. On the contrary, we introduce a new idea which embeds the sampling process inside the similarity evaluation. It automatically omits some voxels at regular spatial intervals from the similarity computation.

\subsection{Approximate Average Common Submatrix}

We compute the similarity between CT exams by using a new method that compares voxel patches where a portion of voxels is omitted at regular intervals. It represents an extension in 3D of the state-of-the-art \emph{Approximate Average Common Submatrix} (A-ACSM)  similarity measure \cite{[15]}, \cite{[16]} used for comparing color or gray level 2D images. For an easier understanding of the approach, we describe the baseline A-ACSM procedure in 2D. Then, we explain the introduced modifications in 3D to the baseline method.

\begin{figure}[t!]
\begin{center}
\includegraphics[height=7cm, width=7cm, keepaspectratio]{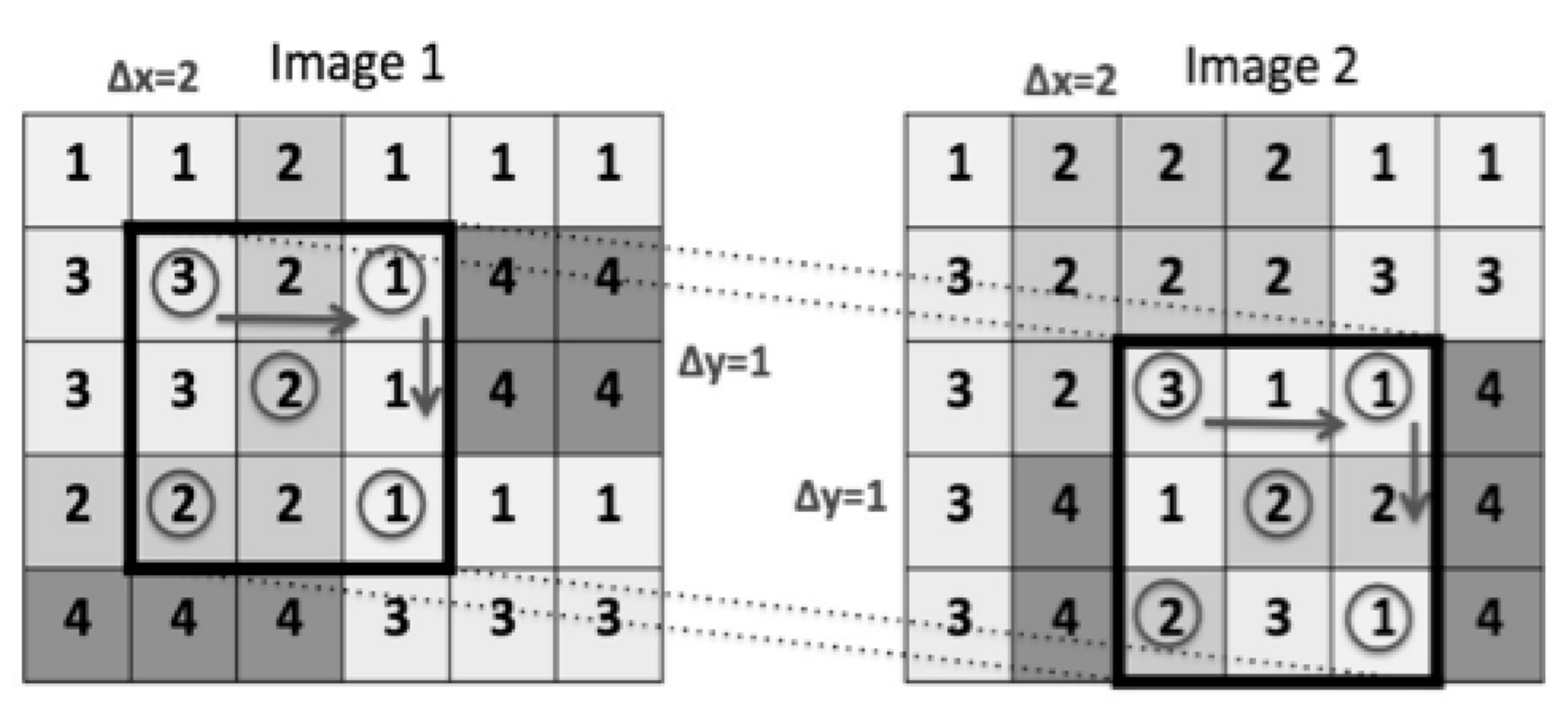}
\caption{A demonstration sample of match between two patches belonging to images 1 and 2. Each image has four gray levels which are also numbered from 1 to 4. An interval of $\Delta x=2$ and $\Delta y=1$ is set respectively along the columns and rows of the patches. Accordingly, the match is only verified between the elements in the circles. The elements are selected as in a chessboard.  In this case, the two image patches perfectly match because all the elements in the circles correspond to one another.}
\label{Fig7}
\end{center}
\end{figure}

\begin{figure}[t!]
\begin{center}
\includegraphics[height=8cm, width=8cm, keepaspectratio]{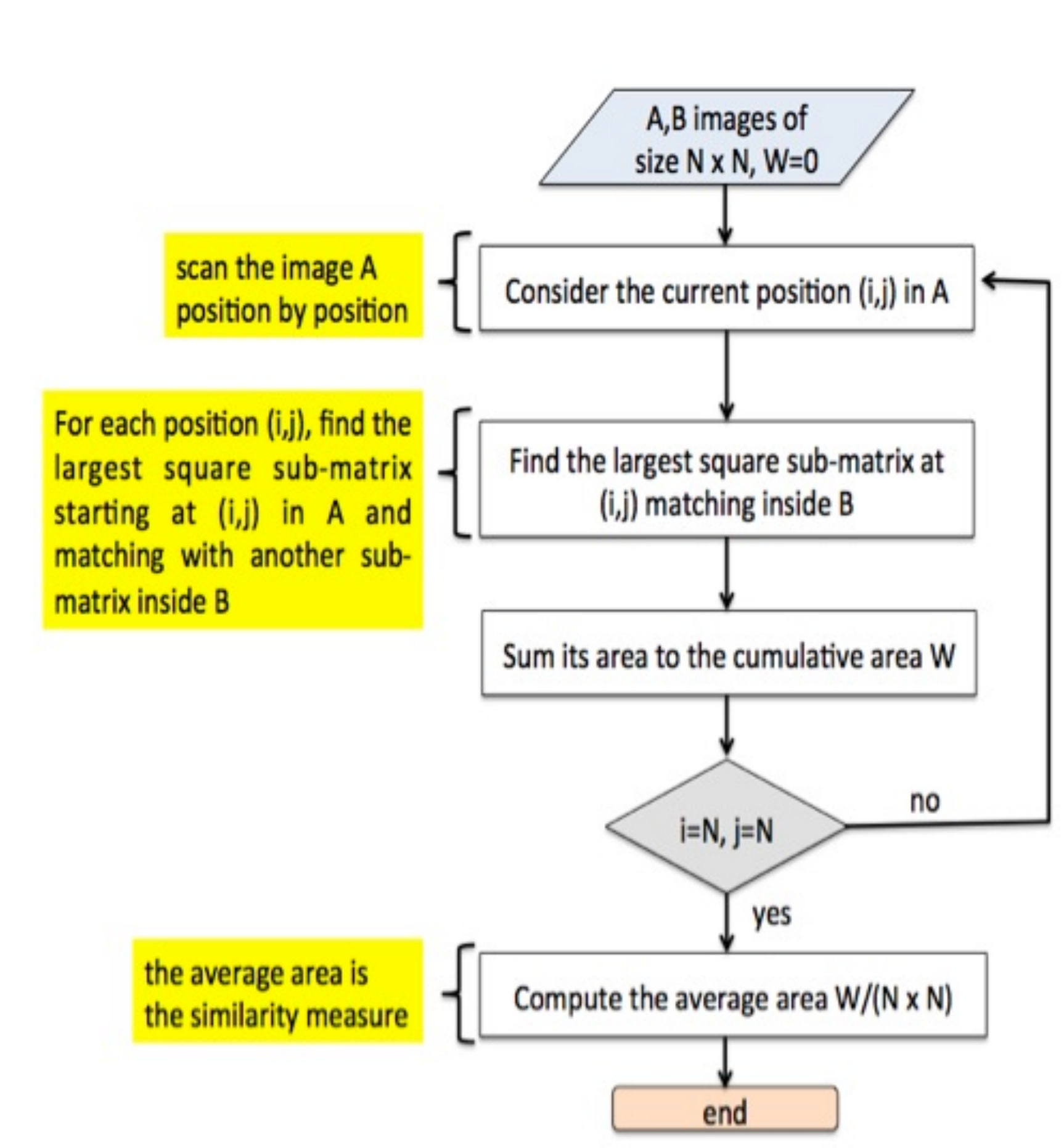}
\caption{Flowchart of the A-ACSM algorithm }
\label{Fig8}
\end{center}
\end{figure}

\begin{figure}[t!]
\begin{center}
\includegraphics[height=7cm, width=7cm, keepaspectratio]{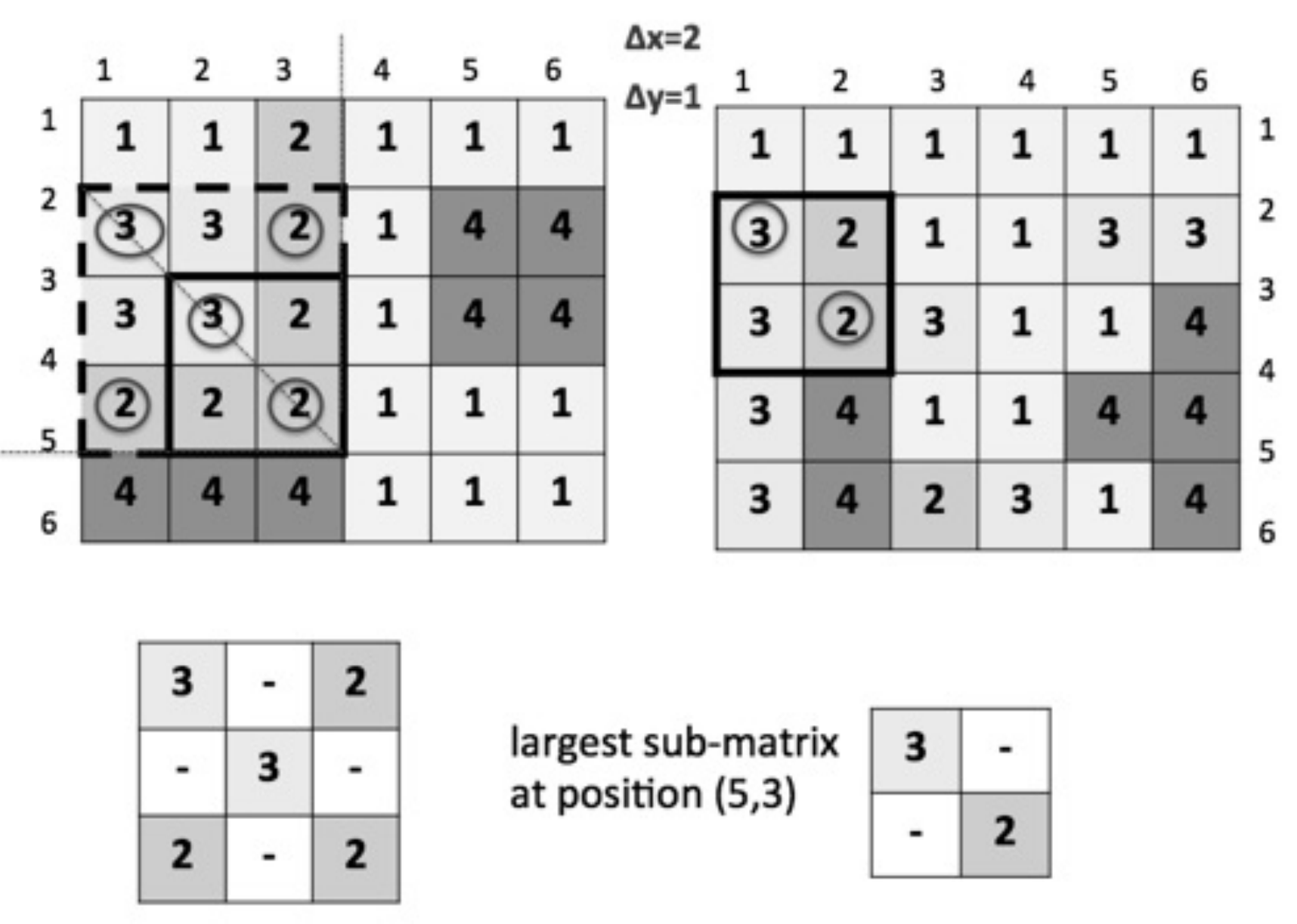}
\caption{The largest square sub-matrix at position (5,3) of the first image. All the sub-matrices of different size along the main diagonal at (5,3) are considered to match inside the second image. In this case, the sub-matrix of size 3 has no match inside the second image. Hence, the sub-matrix of size 2 is considered. Because it has a correspondence at position (3,2) in the second image, it is the largest square sub-matrix at position (5,3).}
\label{Fig9}
\end{center}
\end{figure}

The advantage of A-ACSM is that it does not need to extract complex descriptors from the images to be used for the comparison. On the contrary, an image is considered as a matrix, and the similarity is evaluated by measuring the average area of the largest sub-matrices which the two images have in common. The principle underlying this evaluation is that two images can be considered as similar if they share large patches representing image patterns. Two patches are considered as identical if they match in a portion of the pixels which are extracted at regular intervals along the rows and columns of the patches. Hence, the measure is an easy match between a portion of the pixels. This concept introduces an approximation, which is based on the ``naive" consideration that two images do not need to exactly match in the intensity of every pixel to be considered as similar. This approximation makes the similarity measure more robust to noise, i.e., small variations in the pixels' intensity due to errors in image generation, and considerably reduces its execution time when it is applied on large images, because a portion of the pixels does not need to be checked. Fig. \ref{Fig7} shows a sample of match between two image patches with an interval of two along the columns and one along the rows of the patches. Fig. \ref{Fig8} depicts the algorithm for computing the A-ACSM similarity measure. Fig. \ref{Fig9} shows how to find the largest square sub-matrix at a sample position (5,3) of the first image. 

\subsection{3D Approximate Average Common Submatrix}
The extension of A-ACSM in 3D considers the input CT exams to be compared as parallelepipeds. Consequently, the similarity between two CT exams C1 and C2 is computed as the average volume of the largest sub-cubes which the two CT exams have in common. Two sub-cubes are considered as identical if they match in a portion of the voxels which are extracted at regular intervals along the $x$, $y$ and $z$ axes of the sub-cubes. We also introduce a further modification in the baseline algorithm depicted in Fig. \ref{Fig8} (see step 2). In particular, for each position ($i,j,k$), we search for the largest sub-cube in C1 starting from ($i,j,k$) and matching in a neighbourhood of given volume $v$ from position ($i,j,k$) inside C2, where $v$ is an input parameter. Hence, we verify the match only in a restricted neighborhood inside C2. It increases the sensitivity of the similarity measure to rigid transformations such as translation and rotation. Also, it provides the matching of corresponding or spatially contiguous CT scans in the two CT exams. Finally, it improves the time complexity of the procedure: the largest common sub-cube only needs to be searched in a restricted volume of the second CT exam, and only a subset of voxels in the sub-cube needs to be compared. At the end, we modify the similarity computation by introducing a normalisation step which divides the similarity value by the maximum, for having a range between 0 (no match) and 1 (maximum match). It simplifies the maximisation process of the similarity function during the registration.

\subsection{Image fusion}
After the registration process, the source CT exam and transformed target CT exam are fused into a single output representation, which keeps a higher quality information derived from the input exams. In particular, we use a simple pixel-based fusion which includes the typical arithmetic operator of subtraction, obtaining good results in some contexts of medical imaging \cite{[14]}. It is able to easily visualize the eventual changes occurred over time in the different CT exams.

In particular, the method employs the subtraction operator \cite{[17]} on the source and transformed target CT exams in order to determine a signed difference $D$ as follows \cite{[18]}:
\begin{equation}
D(x, y, z)=A(x, y, z)- \overline{B}(x, y, z),
\end{equation}
where $\overline{B}=T(B)$  derives from the image registration process. Then, $D$ is thresholded for generating a change map $M$ as follows:
\begin{equation}
M(x, y, z)=\begin{cases} 1, & \mbox{if } |D(x, y, z)|>t \\ 0, & \mbox{otherwise} \end{cases},
\end{equation}
where $t$ is the threshold parameter. Finally, the change map $M$ is visualised by the system together with the aligned CT exams.

\section{Expected Results and Benefits}
From an experimental point of view, the system is expected to be more accurate in registering CT exams of stroke than the current state-of-the-art systems. This is mainly due to the use of a common submatrix-based similarity measure, which revealed to overcome the performances of other distance or similarity measures on different types of images. These same measures are currently adopted for CT exam registration (e.g. entropy-based measures, like Mutual Information) \cite{[19]}. Also, because of its robustness to noise due to the embedded sampling strategy, 3D A-ACSM is expected to be a reliable candidate measure for comparing CT exams in image registration.

Differently from the previous methods, the time-based monitoring of the stroke is here in the focus in a period of time which can be established by the physician by selecting the CT exams to compare. It will provide a more complete and customised analysis of the patient's state according to the needs of the physician. 

Finally, the proposed system in its current form represents a novelty in the state-of-the-art and is expected to be primarily of high utility and support for industrial partners as well as hospital structures in Croatia and in the rest of the world. Table \ref{tab1} summarises the main discussed expected results and corresponding benefits derived from the system.

\begin{table*}[t!]
\caption{Expected results and benefits derived from the proposed system}
\begin{center}
\begin{tabular}{|c|c|}
\hline
\bf Expected result	&\bf Benefit\\\hline
Customised time-based monitoring of the stroke&Analysis of the patient's state \\
&according to the needs of the physician\\\hline
New system in the state-of-the-art&	High support for industrial partners \\
&as well as hospital structures \\\hline
Robust to noise similarity measure &	Higher accuracy in \\
for image registration&registering CT exams of stroke\\
\hline
\end{tabular}
\end{center}
\label{tab1}
\end{table*}%

\section{Conclusion and Future Work}
This paper introduced a new system for the temporal monitoring of the stroke using an image registration method followed by image fusion of CT exams. A new similarity measure was also proposed for the comparison of CT exams during the registration phase. It will provide important benefits in terms of lower costs, faster and customised support in the diagnosis process, and higher accuracy and robustness in tracking the manifestation and evolution of the stroke.

Future work will provide results of evaluation and testing of the system.

\end{document}